\documentclass[runningheads]{llncs}
\usepackage[colorinlistoftodos]{todonotes}
\usepackage{amssymb}
\usepackage{stmaryrd}
\usepackage{amsmath}
\usepackage{bm}
\usepackage[skins,theorems]{tcolorbox}
\usepackage{enumitem}
\usepackage{subcaption}
\usepackage{graphicx}
\usepackage{dirtytalk}
\usepackage{geometry}
\usepackage{geometry}
\begin{document}
\title{Regularization Shortcomings for Continual Learning}
%%%%%%%%%%%%%%%%%%%%%%%%%%%%%%%%%%%%%%%%%%%%%%%%%%%%%%%%%%%%%%%%%
%%%%%%%%%%%%%%%%%%%%%%%%%%%%%%%%%%%%%%%%%%%%%%%%%%%%%%%%%%%%%%%%%

%Authors INFOS

%%%%%%%%%%%%%%%%%%%%%%%%%%%%%%%%%%%%%%%%%%%%%%%%%%%%%%%%%%%%%%%%%
%%%%%%%%%%%%%%%%%%%%%%%%%%%%%%%%%%%%%%%%%%%%%%%%%%%%%%%%%%%%%%%%%
\author{Timoth\'ee Lesort *\inst{1} \orcidID{0000-0002-8669-0764} \and
Andrei Stoian\inst{3} \orcidID{0000-0002-3479-9565}  \and
David Filliat\inst{2} \orcidID{0000-0002-5739-1618} }
\authorrunning{T. Lesort et al.}
\institute{ University of Montréal, MILA - Quebec AI Institute \and
Flowers Laboratory (ENSTA ParisTech \& INRIA), France \and
 Thales, Theresis Laboratory, France \\
 * Corresponding author \footnote{timothee.lesort@mila.quebec}}
%%%%%%%%%%%%%%%%%%%%%%%%%%%%%%%%%%%%%%%%%%%%%%%%%%%%%%%%%%%%%%%%%
%%%%%%%%%%%%%%%%%%%%%%%%%%%%%%%%%%%%%%%%%%%%%%%%%%%%%%%%%%%%%%%%%
%%%%%%%%%%%%%%%%%%%%%%%%%%%%%%%%%%%%%%%%%%%%%%%%%%%%%%%%%%%%%%%%%

%\author{}%\and
%Second Author\inst{2,3}\orcidID{1111-2222-3333-4444} \and
%Third Author\inst{3}\orcidID{2222--3333-4444-5555}}
%\institute{Alpha University, Princeton NJ 08544, USA}% \and
%Springer Heidelberg, Tiergartenstr. 17, 69121 Heidelberg, Germany
\maketitle              % typeset the header of the contribution
\begin{abstract}

In most machine learning algorithms, training data is assumed to be independent and identically distributed (iid).  
When it is not the case, the algorithm's performances are challenged, leading to the famous phenomenon of \textit{catastrophic forgetting}. Algorithms dealing with it are gathered in the \say{Continual Learning} research field. In this paper, we study the \textit{regularization} based approaches to continual learning and show that those approaches can not learn to discriminate classes from different tasks in an elemental continual benchmark: the class-incremental scenario.
We make theoretical reasoning to prove this shortcoming and illustrate it with examples and experiments.
Moreover, we show that it can have some important consequences on continual multi-tasks reinforcement learning or in pre-trained models used for continual learning.
We believe that highlighting and understanding the shortcomings of regularization strategies will help us to use them more efficiently. 
%We believe this paper to be the first to propose a theoretical description of regularization shortcomings for continual learning.
%
\end{abstract}
\section{Introduction}
%
% IID assumption
The data iid assumption (identically and independently distributed) supposes all data are randomly sampled from a static distribution. This assumption is mandatory in most machine learning algorithms. Unfortunately, in many cases, the assumption cannot be satisfied.
% CL
Continual Learning is a sub-field of machine learning with non-iid data. Its goal is to learn the global optima to an optimization problem where the data distribution changes over time \cite{Lesort2019Continual}. 
In this paper, we study the case where the data distribution is iid by parts. Each iid part is referred to as a \textit{task} and the data distribution changes are signaled by a task label. This type of continual learning scenario has been widely in the continual learning bibliography \cite{kirkpatrick2017overcoming,rebuffi2017icarl,shin2017continual,lesort2018marginal,prabhu12356gdumb,douillard2020podnet}
Each task contains different classes.
In continual learning, this scenario is called a class-incremental or disjoint-task scenario.
The task label is provided for training but not for inference. This setting is referred to as the \textit{single-head} setting.

In this paper, we study a classical approach for continual learning: regularization. We show that in the scenario of class-incremental tasks, this approach has theoretical limitations and can not be used alone. Indeed, it can not learn to distinguish classes from different tasks (inter-tasks classes).

% plan
In section \ref{sec:related_works}, we will present a quick overview of continual learning strategies and their specificities. Tn section \ref{sec:setting}, we will formalize the scenario and the regularization approach to continual learning. we are studying and then demonstrate the regularization shortcomings in section \ref{sec:shortcomings}. This demonstration is illustrated by simple examples in section \ref{sec:examples} and by experiments in section \ref{sec:experiments}. We also present other contexts where regularization is not suited for continual learning in section \ref{sec:applications} and we discuss and conclude this paper in section \ref{sec:conclusion}.
%
%We believe this paper presents important results for a better understanding of CL which will help practitioners to choose the appropriate approach for practical settings.
%
\section{Related works}
\label{sec:related_works}
In continual learning, algorithms protect knowledge from catastrophic forgetting \cite{French99} by saving them into a memory. The memory should be able to incorporate new knowledge and protect past ones from deterioration.
In continual learning, we distinguish three types of memorization mechanisms: (1) \textit{dynamic architecture}  \cite{Rusu16progressive,Li17learning} where neural network use specific weights for specific tasks and can dynamically create new weights for new tasks, (2) \textit{replay methods} where data from past tasks are saved to be replayed later, either as raw data (rehearsal) \cite{Aljundi2019Online,Belouadah2018DeeSIL,wu2019large,Hou2019Learning,caccia2019online}, or as a model that can regenerate then (generative replay)  \cite{shin2017continual,lesort2018generative,wu2018memory} and (3) \textit{regularization} \cite{kirkpatrick2017overcoming,Zenke17,Ritter18Online,schwarz2018progress,farajtabar2019orthogonal} where the important weights from old tasks are protected from modification.

In continual learning, algorithms are trained on a sequence of tasks and each task can be annotated by a label: the task label.
The task label $t$ (typically a simple integer) is an abstract representation built to help continual algorithms to learn. It is designed to index the current task and notify if the task changes \cite{Lesort2019Continual}. 

The capacity of algorithms to learn depends on the memory mechanisms but also on the availability of the task label. It is important for learning and for inference, in particular at test time.
\textit{Dynamic architecture} is a well-known method that needs the task label at test time for an inference. Indeed, since the weights for inference are different for different tasks, the task test label is needed to choose the right weights \cite{Rusu16progressive,Li17learning}.
Rehearsal and Generative Replay methods generally need the task label at training time but not for inferences \cite{lesort2018generative,lesort2018marginal}.
Finally, Regularization methods are often assumed as methods that need task labels only at training time.
In this article, we show that in class-incremental scenarios, regularization approaches can not learn a correct solution without the task label for inference. Our conclusions on regularization shortcomings are similar to \cite{knoblauch2020optimal} ones but we used a different approach.

Test task labels have been used in many continual learning approaches, in particular in those referred to as \say{multi-heads} \cite{deLange2019continual}.
However, the need for task labels for inferences makes algorithms unable to make autonomous predictions and dependant on external supervision for inference.

%%%%%%%%%%%%%%%%%%%%%%%%%%%%%%%%%%%%%%%%%%%%%%%%%%%%%%%%%%%%%%%%%%%%%%%
%%%%%%%%%%%%%%%%%%%%%%%%%%%%%%%%%%%%%%%%%%%%%%%%%%%%%%%%%%%%%%%%%%%%%%%

%%%%%%%%%%%%%%%%%%%%          NEW SECTION          %%%%%%%%%%%%%%%%%%%%

%%%%%%%%%%%%%%%%%%%%%%%%%%%%%%%%%%%%%%%%%%%%%%%%%%%%%%%%%%%%%%%%%%%%%%%
%%%%%%%%%%%%%%%%%%%%%%%%%%%%%%%%%%%%%%%%%%%%%%%%%%%%%%%%%%%%%%%%%%%%%%%
\section{The Regularization approach}
\label{sec:setting}
 %
%\todo{Section Critique}
%
\subsection{The Class-Incremental Scenario}
% setting
The class-incremental scenario consists of learning sets of classes incrementally. Each task is composed of new classes. As the training ends, the model should classify data from all classes correctly without the task label, i.e. single head setting.
% decompose pb
 In this setting, the model needs to both learn the discrimination of intra-task classes and the inter-task classes discrimination (i.e. distinctions between classes from different tasks). 
On the contrary, if the task label was available for inference (e.g. in a multi-head setting), only the discrimination of intra-task classes would have to be learned. The discrimination upon different tasks would have been given by the task label.

% ccl
Learning without access to task labels at test time is then much more complex since it needs to discriminate data that are not available at the same time in the data stream.

In such settings, we would like to demonstrate that regularization does not help to learn the discrimination between inter-tasks classes. 
% example
For example, if the first task is to discriminate white cats vs black cats and the second is the same with dogs, a regularization-based method does not provide the learning criteria to learn features to distinguish white dogs from white cats.

In the next section, we will introduce our formalism of continual learning and regularization and show how regularization strategies are made to protect knowledge from forgetting.

%%%%%%%%%%%%%%%%%%%%%%%%%%%%%%%%%%%%%%%%%  Subsection 
\subsection{Formalism}

In this paper, we assume that the data stream is composed of $N$ disjoint tasks learned sequentially one by one (with $N>=2$).
Task $t$ is noted $T_t$ and we note $x,y \sim T_t$ the data sampled from task $t$ with $x$ the data point and $y$ its label. 
The task label $t$ is a simple integer indicating the task index.
We refer to a sequence of tasks until task $t$ as the continuum $C_t$, with $C_N$ the full sequence of tasks.
At time $t$, the algorithm has access to data from $T_{t}$ only.

We study a disjoint set of classification tasks where classes of each task only appear in this task and never again.
We assume at least two classes per task (otherwise a classifier cannot learn).

 Let $f$ be a function parametrized by $\bm{\theta}$ that implement the neural network's model.
At each task $t$ the model learn an optimal set of parameters $\bm{\theta}^*_t$ optimizing the task loss $\ell_{t}(\cdot)$.
Since we are in a continual learning scenario, $\bm{\theta}^{\ast}_t$ should also be an optima for $C_{t}$, i.e., for all past tasks. 
An optima $\bm{\theta}_t^\ast$ for $C_t$ 
 is a set of parameters which at test time will, for any data point $x$ from $C_t$, classify correctly without knowing from which task $x$ is coming from. 
Therefore, the loss to optimize when learning a given task $t$ is augmented  with a remembering loss:
\begin{equation}
\ell_{C_t}(f(x; \bm{\theta}), y) = \ell_{t}(f(x; \bm{\theta}), y) + \lambda \Omega(C_{t-1})
\label{eq:continual_loss}
\end{equation}
%
%Description
where $\ell_{C_t}(.)$ is the continual loss, $\ell_{t}(.)$ is the current task loss, $\Omega(C_{t-1})$ is the remembering loss on $C_{t-1}$, $\lambda$ is the importance parameter of the remembering loss.

In continual learning, the regularization approach is to define $\Omega(\cdot)$ as a regularization term to maintain knowledge from $C_{t-1}$ 
in the parameters $\bm{\theta}$.
To do so, the regularization loss is designed to penalize the modification of important weights and encourage the use of other weights to learn a new task. 

%
% regu term
The regularization term $\Omega_{t-1}$ act as a memory of $\bm{\theta}^{\ast}_{t-1}$. 
This memory term depends on the learned parameters $\bm{\theta}^{\ast}_{t-1}$, on $\ell_{C_{t-1}}$ the loss computed on $T_{t-1}$ and the current parameters $\bm{\theta}$. 
$\Omega_{t-1}$ memorizes the optimal state of the model at $T_{t-1}$ and generally the importance of each parameter with regard to the loss $\ell_{C_{t-1}}$. 
% all regularization terms
We note $\Omega_{C_{t-1}}$ the regularization term memorizing optimal parameters for all past tasks. One specificity that we want to highlight is that the regularization terms here described do not use saved past data to regularize current loss and only penalized the modification of weights useful in the past.

Many regularization strategies such as \cite{kirkpatrick2017overcoming,Ritter18Online,Zenke17} strictly follow the formalism of eq. \ref{eq:continual_loss}. Other methods such as \cite{farajtabar2019orthogonal,doan2020theoretical} are a bit different, they do not add a regularization loss but they regularize the gradient of the loss to protect important weights. This difference does not impact the remaining of the paper, both approaches are designed to protect important weights without saving data.

%%%%%%%%%%%%%%%%%%%%%%%%%%%%%%%%%%%%%%%%%%%%%%%%%%%%%%%%%%%%%%%%%%%%%%%
%%%%%%%%%%%%%%%%%%%%%%%%%%%%%%%%%%%%%%%%%%%%%%%%%%%%%%%%%%%%%%%%%%%%%%%

%%%%%%%%%%%%%%%%%%%%          NEW SECTION          %%%%%%%%%%%%%%%%%%%%

%%%%%%%%%%%%%%%%%%%%%%%%%%%%%%%%%%%%%%%%%%%%%%%%%%%%%%%%%%%%%%%%%%%%%%%
%%%%%%%%%%%%%%%%%%%%%%%%%%%%%%%%%%%%%%%%%%%%%%%%%%%%%%%%%%%%%%%%%%%%%%%
\section{Regularization Shortcomings for Continual Learning}
\label{sec:shortcomings}
In this section, we will prove that regularization strategies can not ensure continual learning in class-incremental scenarios.
%
%\todo{Section Critique}
%
\subsection{Demonstration}
\label{sub:demo}

We consider in this demonstration the two first tasks of a scenario $T_0$ and $T_1$ and show that we can not learn to discriminate $T_0$ classes from $T_1$ classes.

Let $\{Y_0\}$ and $\{Y_1\}$ be the sets of classes at task $T_0$ and $T_1$ respectively, with  $\{Y_0\} \cap \{Y_1\} = \emptyset$. We will write $(x,y)\sim T_0$ for data sampled from task 0, with $x$ the data point and $y$ the label.

We have a neural network that should be trained to realize the following operation:
\begin{equation}
    \begin{split}
\forall (x,y) \sim (T_0 \cup T_1)\\
y = argmax ( A \cdot \phi(x; \bm{\theta}))
\label{eq:base}
\end{split}
\end{equation}
With $\phi(\cdot;\bm{\theta})$ the feature extractor parametrized by $\bm{\theta}$ and $A$ is the output layer matrix of size $h \times C$ with $h$ the latent space size and $C$ the number of classes $Card(\{Y_0\})+Card(\{Y_1\})$. 

In continual learning, we want to train sequentially on $T_0$ and $T_1$ to learn $\bm{\theta}^*$ and $A^*$ such as eq. \ref{eq:base} is respected.

For the purpose of the demonstration, we rewrite eq. \ref{eq:base} as:
\begin{equation}
    \begin{split}
& \forall (y_i,y_j) \in \{Y_0\}\cup \{Y_1\}, \forall (x, y) \sim (T_0 \cup T_1),\\
& y=y_i \xrightarrow[]{} \langle \phi(x;\bm{\theta}), A^*_i\rangle  > \langle \phi(x, \bm{\theta}), A^*_j\rangle
\end{split}
    \label{eq:objective}
\end{equation}
With $A^*_i$ the i-eme vector of matrix $A^*$ corresponding to class $y_i$ and $\langle \cdot, \cdot \rangle$ the scalar product.

%We can note that to ensure that the proper $\bm{\theta}^*$ and $A^*$ are learned, we must be able to compute $\phi(x;\bm{\theta})$, therefore, we need access to $x$. \david{On dirait que c'est déjà la conclusion qui dit qu'on ne peut pas apprendre ??? Ou alors tu veux dire autre chose, juste qu'on est pas capable de le vérifier même si on l'avait appris  ? Précise.}

%We suppose that the feature of $T_0$ are not sufficient to solve $T_1$, then $A$ and $\phi(\cdot;\bm{\theta})$ needs to be learned during both tasks.
%
%
\medskip
\textbf{- Learning $T_0$}

While learning $T_0$, we do not know anything about $T_1$. Training on $T_0$, is then a classical iid training that learn a set of parameter $\bm{\theta}^*_0$ and an output layer $A^*_{0}$ of size $h \times Card(\{Y_0\})$ such as:
\begin{equation}
    \begin{split}
& \forall (y_i,y_j) \in \{Y_0\}, \forall (x, y) \sim T_0,\\
& y=y_i \xrightarrow[]{} \langle \phi(x;\bm{\theta}^*_0), A^*_{0:i}\rangle  > \langle \phi(x, \bm{\theta}^*_0), A^*_{0:j}\rangle
\end{split}
    \label{eq:objective_T0}
\end{equation}
\medskip
\textbf{- Computing Regularization term}

% ---> cette hypothese est expliquée plus tot dans le papier <----
We compute a regularization term $\Omega_0$ that, we assume, ensures that eq. \ref{eq:objective_T0} is preserved.

The regularizer is designed to preserve the inequality from eq. \ref{eq:objective_T0} but the weights $\bm{\theta}^*_0$ need to be updated to solve the next task and hence $\phi(\cdot;\bm{\theta})$  will evolve (as well as the matrix $A$).

\medskip

\textbf{ - Solving eq. \ref{eq:objective} while Learning $T_1$ with regularization}

At task $T_1$ we do not have access to data from $T_0$ anymore and the model have to be trained such as $\bm{\theta}^*_1$ and $A^*_1$ are a set of weights that ensures eq. \ref{eq:objective}. We add to $A^*_0$ several column to be able to predict the new classes of $T_1$. The dimensions of $A$ are now $h \times (Card(\{Y_0\})+Card(\{Y_1\}))$.

We will show that it is impossible to verify eq. \ref{eq:objective} in such setting (i.e., with eq. \ref{eq:objective_T0} guaranty but without access to any data of $T_0$).
Indeed, in order to verify eq. \ref{eq:objective} we need notably (but not exclusively) to ensure that:
\begin{equation}
    \begin{split}
& \forall (x_i, y_i) \sim T_0, \forall y_j \sim \{Y_1\} \\
& \langle \phi(x_i;\bm{\theta}^*_1), A^*_{1:i}\rangle  > \langle \phi(x_i; \bm{\theta}^*_1), A^*_{1:j}\rangle \\
\end{split}
    \label{eq:pb_T1}
\end{equation}
The principal problem here is that at task 1 we do not have access to $x_i$ since it is a data point from $T_0$. Hence, we can not estimate $\langle \phi(x_i;\bm{\theta}^*_1), A^*_{1:j}\rangle$ and we can not ensure the inequality \ref{eq:pb_T1}.
Moreover, we could not estimate this product at task 0 since $A^*_{1:j}$ was not created yet. 

Inequality \ref{eq:pb_T1} can not be optimized and we can not know if it is true or not. Therefore, if inequality \ref{eq:pb_T1} is not already true when task T1 starts, the learning algorithms can not change it. 

\medskip

\textbf{- Conclusion}

The regularization strategy ensures inequalities from eq.  \ref{eq:objective_T0} but not inequalities from eq. \ref{eq:pb_T1}. Therefore, we can not estimate if $A^*_1$ and $\bm{\theta}^*_1$ are suited to verify eq. \ref{eq:base}.

%The demonstration was based on the two first tasks in a scenario but the same reasoning can directly be applied to any sequential two tasks.

We can deduce from this demonstration that protecting weights from past tasks, as regularization strategies do, is not sufficient to ensure continual learning in class-incremental scenarios.

%Hence, we showed that the regularization strategy is not able to learn continually 
%The demonstration was based on the two first tasks in a scenario but the same reasoning can directly be applied to any sequential two tasks.

%This problem proves that using a regularization strategy only is not sufficient to solve a class-incremental scenario in continual learning.

\subsection{The Feature Selection Problem}
\label{sub:feature_selection}

The inequality \ref{eq:pb_T1} can typically be false if images from class $y_i$ and $y_j$ have some common features. 
While learning on a single task, the learning algorithm does not have access to the full data distribution. Then, it might not have enough information to select and learn features useful to solve the full scenario. The learning algorithm might rely on \say{bad features} to solve the current task. This continual learning problem can be labeled as a \textit{feature selection problem}.
Hence, discriminative features for the global problem can then be considered locally as non-discriminative features or the opposite.

Let go back to inequality \ref{eq:pb_T1}. Let $a_{ij}$ be some common features of $y_i \in T_0$ and $y_j \in T_1$. 
$a_{ij}$ features might be ignored because not discriminative at tasks 0. But at task 1, they are considered as discriminative features. They are learned by the feature extractor and used by the output layer $A^*_{1:j}$ for class prediction. Then, any input images with $a_{ij}$ features can be potentially classified as class $j$, notably $y_i$ images. Hence, inequality \ref{eq:pb_T1} risk to be false and the model will be unable to detect it. 
As noted by \cite{doan2020theoretical,ramasesh2021anatomy} when tasks have similarities (e.g. common features) the model is more prone to forget. We argue it might be due to this feature selection problem

\subsection{Shortcomings in the Feature Extractor}
\label{sub:features}

In section \ref{sub:demo}, we proved that the regularization strategies can not ensure continual learning with a model composed of a feature extractor and a linear output layer in a class-incremental scenario. In section \ref{sub:feature_selection}, we explained that the model can not spontaneously select the right features while training on a task.

%The demonstration highlights a problem in the output layer and one could imagine that another output layer could solve the problem. 
In this section, we would give some insight into more general problems in the feature extractor that make continual learning challenging for regularization approaches.
%that would make this solution inefficient.

In our model, the feature extractor's goal is to learn a projection of input data into space where they are linearly separable or at least where the output layer can learn a solution to the classification problem. 
Hence, the feature extractor needs to learn discriminative features that make it possible to map different classes into different embedding representations.

Unfortunately, to learn discriminative features between two different classes the feature extractor needs to have access to both of them simultaneously. %If the model is pretrained and has already the rights features to discriminates those classes, there is no problem but 
The model can trivially not learn new features to discriminate a class from an inaccessible other one. The regularization strategy does not provide access to data from past tasks, then the feature extractor can not learn discriminative features to tell apart past data from current ones.

Therefore, the feature extractor will not be able to learn features that discriminate data from different tasks which might lead to interference between tasks in the latent space.

\subsection{Summary}

In this section, we showed that the regularization-based methods can not ensure continual learning in class-incremental scenarios. Their memorization mechanisms do not make it possible to ensure discrimination between inter-tasks classes.

%%%%%%%%%%%%%%%%%%%%%%%%%%%%%%%%%%%%%%%%%%%%%%%%%%%%%%%%%%%%%%%%%%%%%%%
%%%%%%%%%%%%%%%%%%%%%%%%%%%%%%%%%%%%%%%%%%%%%%%%%%%%%%%%%%%%%%%%%%%%%%%

%%%%%%%%%%%%%%%%%%%%          NEW SECTION          %%%%%%%%%%%%%%%%%%%%

%%%%%%%%%%%%%%%%%%%%%%%%%%%%%%%%%%%%%%%%%%%%%%%%%%%%%%%%%%%%%%%%%%%%%%%
%%%%%%%%%%%%%%%%%%%%%%%%%%%%%%%%%%%%%%%%%%%%%%%%%%%%%%%%%%%%%%%%%%%%%%%

\section{Simple Examples}
\label{sec:examples}

\subsection{Minimal Example}

In this section, we will illustrate by a minimal example that even with two tasks of binary classification, the protection of important weights by regularization (or by anything else) is not enough to learn continually in a class-incremental scenario. This example is also representative of the features selection problem discussed in section \ref{sub:feature_selection}.

\medskip

\textbf{First task:}
We have two different data points for the task $T_0$

($X0=[-1,1]$, $y=0$) and 
($X1=[1,1]$, $y=1$)

We train a linear layer $A$ to solve the binary classification task.

$A_0=[-1,0]$ and
$A_1=[1,0]$

$$\langle X0, A_0 \rangle = 1 > \langle X0, A_1 \rangle = -1$$ 
$$\langle X1, A_0 \rangle = -1 < \langle X1, A_1 \rangle = 1$$

$A=[A_0,A_1]$ is a solution for the classification. We regularize by keeping $A_0$ and $A_1$ unchanged.

\medskip

\textbf{Second task:}

We have two new data points for the task $T_1$

($X2=[0,-2]$, $y=2$) and ($X3=[-3,-2]$, $y=3$)

We train two new vectors in $A$ to solve the binary classification task and we found:

$A_2=[1,2]$ and $A_3=[-1,3]$

$$\langle X2, A_2 \rangle = -4 > \langle X2, A_3 \rangle = -6$$
$$\langle X3, A_2 \rangle = -7 < \langle X3, A_3 \rangle = -3$$

So everything is fine for the second task classification and the first solution was protected since $A_0$ and $A_1$ stayed unchanged.

\medskip

\textbf{Problem:}

$$\langle X0, A_0 \rangle = 1 < \langle X0, A_3 \rangle = 4$$

The classifier would classify $X_0$ to $y=3$ rather than to $y=0$. 
$A$ does not allow the model to classify the $X0$ data point correctly anymore even if we protected the solution of task 0 while learning the solution for task 1.
It was indeed impossible to predict that $\langle X0, A_3 \rangle = 4$ since at $T_0$, $A_3$ did not exists and at $T_1$, $X0$ was not accessible anymore.

This simple example aims to show that protecting past weights is not sufficient to learn continually a class-incremental sequence of tasks. The features selected to solve task 1 where finally not pertinent for the full scenario.

\subsection{Visual Examples}
\label{sub:visual_examples}

To illustrate the shortcomings of regularization approaches proposition from section \ref{sec:shortcomings}, we present two insightful examples of regularization limitations.

\medskip

\textbf{- The Output Layer Problem: }

In the demonstration proof in section \ref{sub:demo}, we showed that there was a problem to learn how to discriminate inter-tasks classes with a regularization strategy.

\begin{figure}[h]
    \centering
    \begin{subfigure}[t]{0.49\linewidth}
        \centering
        \includegraphics[width=0.45\linewidth]{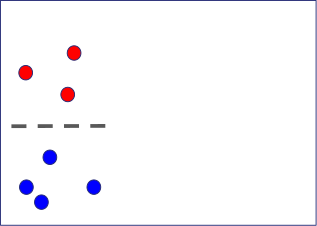}
    \end{subfigure}
    \begin{subfigure}[t]{0.49\linewidth}
        \centering
        \includegraphics[width=0.45\linewidth]{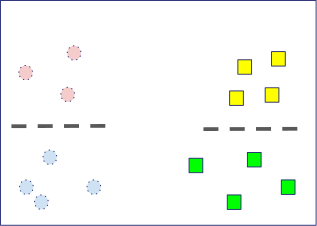}
    \end{subfigure}
    \caption[Simple case of continual learning classification in a multi-task scenario.]{Simple case of continual learning classification in a class-incremental scenario. Left, the task T0: learning a hyper-plane splitting two classes (red and blue dots). Right, the task  T1: learning a line splitting two classes (yellow and green squares) while remembering $T_0$ models without remembering $T_0$ data (pale red and blue dots).}
    \label{fig:simple_continual}
\end{figure}

In this example, we assume that the classes are already linearly separable, i.e. we have a perfect feature extractor that projects data in a 2D plane with each class linearly separable from the others.
In the first task $T_0$, we have two classes. The model learns one hyper-plane $\mathcal{Q}_0$ separating the instances of these two classes (See Figure \ref{fig:simple_continual}). 
In the second task $T_1$, we have two new classes and $\mathcal{Q}_0$ is protected from modification. Then, we learn a hyper-plane $\mathcal{Q}_1$ that separates our two new classes.
In the end, we have learned the hyper-planes $\mathcal{Q}_0$ and $\mathcal{Q}_1$ to distinguish classes from $T_0$ and classes from $T_1$. But none of those hyper-planes helps to discriminate $T_0$ classes from $T_1$ classes, as illustrated Figure \ref{fig:simple_continual}. 

This example shows the limitation of only protecting past weights in a class-incremental scenario even if the classes are linearly separable.

\medskip

\textbf{- The Latent Features Problem: }

In this second example, the feature extractor needs to be updated to learn new features.

\begin{figure}
    \begin{subfigure}[t]{0.49\linewidth}
        \centering
        \includegraphics[width=0.45\linewidth]{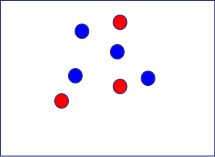}
    \end{subfigure}%
    ~ 
    \begin{subfigure}[t]{0.49\linewidth}
        \centering
        \includegraphics[width=0.45\linewidth]{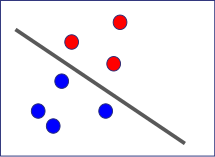}
    \end{subfigure}
    \caption{$T_0$ feature space before learning $T_0$ (Left), $T_0$ feature space after learning $T_0$ with a possible decision boundary (Right). Data points are shown by blue and red dots. The line (right part) is the model learned to separate data into the feature space.}
    \label{fig:T0}
\end{figure}

If we have only two classes in the first task, the model will learn to separate classes instances into two groups with the features extractor $\phi(\cdot,\bm{\theta}^*_0)$ and one hyper-plan $\mathcal{Q}_0$ separating the two classes instances (See Figure \ref{fig:T0}). 

\begin{figure}
        \begin{subfigure}[t]{0.49\linewidth}
        \centering
        \includegraphics[width=0.45\linewidth]{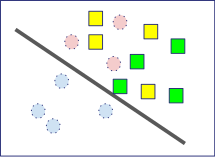}
    \end{subfigure}%
    ~ 
    \begin{subfigure}[t]{0.49\linewidth}
        \centering
        \includegraphics[width=0.45\linewidth]{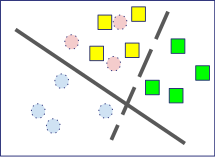}
    \end{subfigure}
    \caption[Case of representation overlapping while continual learning classification in a multi-task scenario.]{Case of representation overlapping while continual learning classification in a class-incremental scenario. At task $T_1$, feature space of $T_1$ before learning $T_1$ (Left), Feature space of $T_1$ after learning $T_1$ with a possible decision boundary (Right). New data are plotted as yellow and green squares and old data that are not available anymore to learn are shown with pale red and blue dots. } 
    \label{fig:T1}
\end{figure}

For the second task, we have two new classes and a regularization protecting $\mathcal{Q}_0$ and $\phi(\cdot,\bm{\theta}^*_0)$. Then, we can learn a features extractor $\phi(\cdot,\bm{\theta}^*_1)$ to disentangle new class instances in the latent space and a hyper-plane $\mathcal{Q}_1$ that separates them.
In the end, we can disentangle classes from $T_0$ and classes from $T_1$ and we have two hyper-planes $\mathcal{Q}_0$ and $\mathcal{Q}_1$ to distinguish classes from $T_0$ and classes from $T_1$. 
But we can not disentangle  $T_0$ classes from $T_1$ classes and none of the learned hyper-planes helps to discriminate $T_0$ classes from $T_1$ classes (See Fig.~ \ref{fig:T1}). It leads to errors in the neural network predictions.
At test time, it will not be possible for the model to discriminate between classes correctly. It illustrates the feature extractor problem mentioned in section \ref{sec:shortcomings}.
However, with the task label for inference, we could potentially perfectly use $\phi(\cdot,\bm{\theta}^*_0)$, $\phi(\cdot,\bm{\theta}^*_1)$, $\mathcal{Q}_0$ and $\mathcal{Q}_1$ to make good predictions.
%The task label at test time allows solving the global problem with only partial solutions.
Nevertheless, assuming that the task label is available for prediction is a strong assumption in continual learning and involves a need of supervision at test time. 
%
% %%%%%%%%%%%%%%%%%%%%%%%%%%%%%%%%%%%%%%%%%%%%%%%%%%%%%%%%%%%%%%%%%%%%%%%
% %%%%%%%%%%%%%%%%%%%%%%%%%%%%%%%%%%%%%%%%%%%%%%%%%%%%%%%%%%%%%%%%%%%%%%%

% %%%%%%%%%%%%%%%%%%%%          NEW SECTION          %%%%%%%%%%%%%%%%%%%%

% %%%%%%%%%%%%%%%%%%%%%%%%%%%%%%%%%%%%%%%%%%%%%%%%%%%%%%%%%%%%%%%%%%%%%%%
% %%%%%%%%%%%%%%%%%%%%%%%%%%%%%%%%%%%%%%%%%%%%%%%%%%%%%%%%%%%%%%%%%%%%%%%
\section{Experiments}
\label{sec:experiments}
%
%\todo{Section Critique}
%
%
In this section, we would like to support the results from section \ref{sec:shortcomings} and show that regularization methods fail to discriminate inter-tasks classes (i.e. classes from different tasks).
We propose two experiments, one with a multi-head output layer and another one with a single head output layer. The goal is to show that regularization methods depend on the task label at test time to succeed.

%need to learn to discriminate data even if they succeed to protect past weights.
%with the same regularization approaches one with a multi-head output layer (using task labels) and another with a single head layer (without task label then). 
%The goal is to show that regularization approaches protect past weights but are inefficient to learn to differentiate classes from different tasks.

\subsection{Scenario and Settings}

\begin{figure}
    \centering
    \begin{subfigure}[t]{0.25\linewidth}
        \centering
        \includegraphics[width=\linewidth]{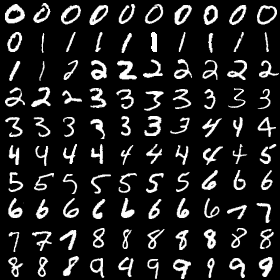}
        \caption{Task 0}
    \end{subfigure}%
    ~ 
    \begin{subfigure}[t]{0.25\linewidth}
        \centering
        \includegraphics[width=\linewidth]{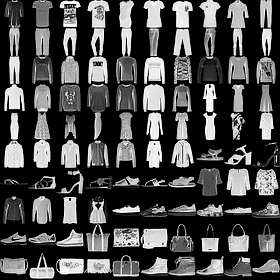}
        \caption{Task 1}
    \end{subfigure}%
    ~ 
    \begin{subfigure}[t]{0.25\linewidth}
        \centering
        \includegraphics[width=\linewidth]{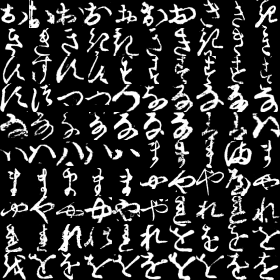}
        \caption{Task 2}
    \end{subfigure}
    \caption[Illustration of the MNIST-Fellowship dataset's three tasks.]{The three 10 classes-classification tasks of the MNIST-Fellowship scenario.}
    \label{fig:MNIST_Fellowship}
\end{figure}

We experiment with the \say{MNIST-Fellowship} scenario proposed in \cite{douillard2021continuum}. This dataset is composed of three datasets (Fig. \ref{fig:MNIST_Fellowship}): MNIST \cite{LeCun10}, Fashion-MNIST \cite{Xiao2017} and KMNIST \cite{clanuwat2018deep}, each composed of 10 classes. The algorithms are sequentially trained on each datasets, i.e. there are three tasks of ten classes. 
The evaluation set is composed of the concatenation of the three datasets' test sets.
%We choose this scenario because it gathers three easy datasets for prototyping machine learning algorithms.
%
As mentioned in the introduction of this section, our goal is to illustrate the limitation of regularization-based methods in class-incremental scenarios, i.e. their inability to learn to discriminate data from different tasks. 
We designed two settings to highlights this phenomenon:
%In particular that they can not distinguish classes from different tasks.  Thus, we propose two different settings one with a multi-head layer, one with a single head layer. %Both are quite similar the only difference is the availability  of the task label for predictions:

 \begin{itemize}
    \item \textbf{1. Single-Head}: In this setting, the task label can not be used for inference. At the test time, data points should be classified among 30 classes without any supplementary information.
    
     \item \textbf{2. Multi-Heads}: In this setting, the task label can be used for inference. Then, the neural network has a specific \say{head} (output layer) for each task. At the test time, data points are annotated with their task label. The task label makes it possible to select a head among the three heads. Once the head is selected, the neural network classifies the data point among 10 classes.
\end{itemize}

One of the main differences between those two scenarios is that in \textit{Single-Head} the neural network has to discriminate inter-tasks classes while in \textit{Multi-Heads} the neural network only has to discriminate intra-task classes. 

In section \ref{sec:shortcomings}, we showed that this inter-tasks  discrimination problem could not be solved by regularization strategies. Then, in this experiment, we will illustrate this limitation (\textit{Single-Head} setting). However, When the inter-task discrimination  does not have to be learned (\textit{Multi-Heads} setting), we will show that the regularization strategies are significantly better. 

We experiment with the most famous regularization strategy EWC \cite{kirkpatrick2017overcoming}, its best working variant EWC-KFAC \cite{Ritter18Online} and a recent regularization strategy, orthogonal gradient descent (OGD) \cite{farajtabar2019orthogonal} which is one of the states of the art regularization strategy in multi-head settings.
We also tested a vanilla rehearsal strategy and a vanilla fine-tuning strategy as baselines.

In our implementation, we used an importance parameter of $1000$ for Ewc and $10$ for Ewc-Kfac, they have been chosen empirically to maximize performance on multi-heads setting and transferred after to single-head setting. For rehearsal, we saved 100 samples per class (1000 per task), and while learning new classes we sampled with the same probability new classes and past classes. 
All algorithms have been trained by stochastic gradient descent with a learning rate of $2e-3$.

\subsection{Results}

\begin{figure}[h]
    \begin{subfigure}[t]{0.5\linewidth}
    \centering
    \includegraphics[width=\linewidth]{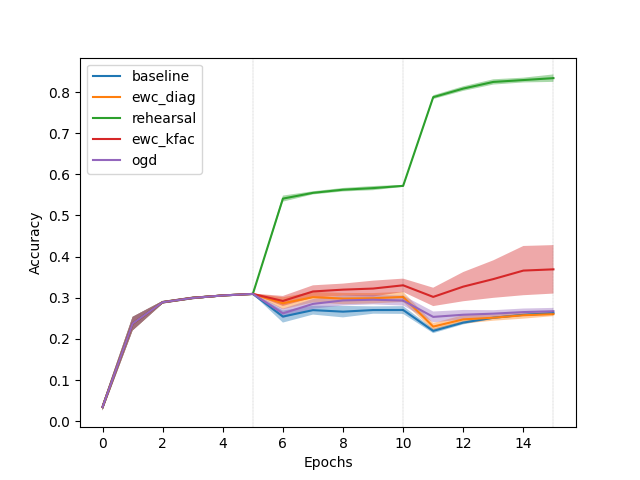}
    \caption[]{Single-Head Experiment}
    \label{fig:SingleH}
    \end{subfigure}
     ~ 
    \begin{subfigure}[t]{0.5\linewidth}
        \centering
        \includegraphics[width=\linewidth]{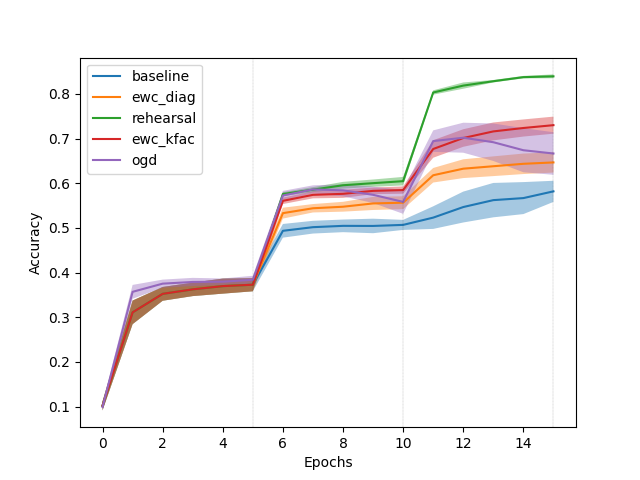}
        \caption[]{Multi-Heads Experiment}
        \label{fig:MultiH}
    \end{subfigure}
    \caption[]{Evolution of the test accuracy of several continual learning approaches on MNIST-Fellowship scenario. The test accuracy is computed on the concatenation of the test sets of all tasks. There are 5 epochs per tasks and task transitions are marked as pale grey vertical lines.}
    \label{fig:results}
\end{figure}

We present the evolution of the test accuracy (fig. \ref{fig:results}) as a result of this experiment.
If we look at the \textit{Single-Head} setting (fig. \ref{fig:SingleH}), i.e. without task label for inference, we can see that the regularization approaches are ineffective. 
However, in the \textit{Multi-Heads} setting (fig. \ref{fig:MultiH}), i.e. when the task label is available, we can see that the regularization strategies perform significantly better than the fine-tuning baseline.

The relative success of regularization strategies in the \textit{Multi-Heads} setting tends to prove that they are effective in their role of protecting weights from past tasks. However, their failures in \textit{Single-Head} setting tend to prove that weight protecting is not sufficient when the task label is not available, i.e. the model can not distinguish clearly inter-tasks classes.

%\medskip

The \say{Multi-Heads} experiment shows us that the regularization strategy has a significant effect on the results which means that the weight protecting role is working. On another hand, in the \say{Single-Head} experiment, the regularization strategies are almost as good as the fine-tuning strategy. We can deduce that the weight protecting role of regularization does not make it possible to replace the role of the task label, i.e. distinguishing class from different tasks. 

In figure \ref{fig:SingleH}, we can see that the Ewc-Kfac works somehow better than the other. However, this improvement seems unstable and still very far from a simple baseline as rehearsal which in both settings is well-performing. Using rehearsal appears to be a good alternative solution to the task labels for regularization strategies in class-incremental scenarios.

Those results support the theoretical shortcomings of regularization strategies described in section \ref{sec:shortcomings}.

\begin{figure}[h]
    \centering
    \includegraphics[width=\linewidth]{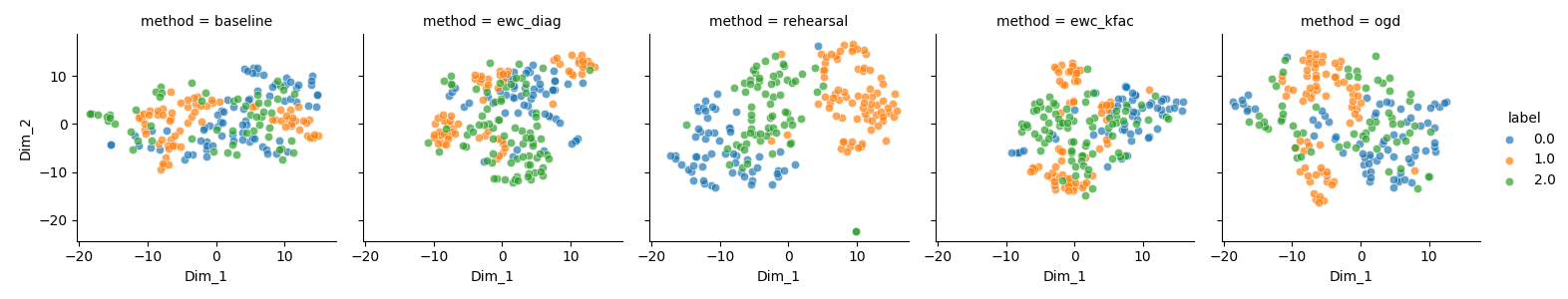}
    \caption[]{2D representation of latent space in the \textit{Single-Head} experiment using t-SNE. The labels are the task labels. The goal of the figure is to visualize how tasks are disentangled by the feature extractor for each strategy.}
    \label{fig:tsne}
\end{figure}

As a supplementary result, we proposed a visualization figure of the neural network's latent space (fig. \ref{fig:tsne}). For this figure, we select 200 images randomly in the test set. Then, for each strategy, we used the trained neural network to compute the latent vectors (the input of the last layer). Then, we apply t-SNE visualization method\footnote{We used the scikit-learn \cite{pedregosa2011scikit-learn} library with their default parameters.
} \cite{Maaten08visualizingdata} on those vectors to get a 2D representation. 
This figure represents the neural network's ability to disentangle data from different tasks. Each task is represented as a color.
The figure suggests that only the latent space of the rehearsal strategy disentangle clearly the data from different tasks. It strengthens the reasoning proposed in section \ref{sub:features} about regularization shortcomings in the feature extractor. %The representation learned seems unable to correctly disentangle data from different tasks in regularization strategies.
%
%The results we present a visualization that illustrates that the latent layer in regularization is not suited to distinguish classes from different tasks.
%
%
% \begin{figure}[h]
%     \centering
%     \includegraphics[width=\linewidth]{New_Figures/comparative_accuracies_per_class.png}
%     \caption[]{Visualizing forgetting with accuracy Per Class}
%     \label{fig:tsne}
% \end{figure}

\medskip
Those experiments highlight the dependence of regularization strategies on the task label to discriminate inter-tasks classes. We also proposed a visualization showing the inability of the representation learned to disentangle tasks data. The results presented endorse the theoretical results from section \ref{sec:shortcomings}.

%%%%%%%%%%%%%%%%%%%%%%%%%%%%%%%%%%%%%%%%%%%%%%%%%%%%%%%%%%%%%%%%%%%%%%%
%%%%%%%%%%%%%%%%%%%%%%%%%%%%%%%%%%%%%%%%%%%%%%%%%%%%%%%%%%%%%%%%%%%%%%%

%%%%%%%%%%%%%%%%%%%%          NEW SECTION          %%%%%%%%%%%%%%%%%%%%

%%%%%%%%%%%%%%%%%%%%%%%%%%%%%%%%%%%%%%%%%%%%%%%%%%%%%%%%%%%%%%%%%%%%%%%
%%%%%%%%%%%%%%%%%%%%%%%%%%%%%%%%%%%%%%%%%%%%%%%%%%%%%%%%%%%%%%%%%%%%%%%

\section{Applications}
\label{sec:applications}

In this section, we point out supplementary shortcomings of regularization in other types of learning situations, namely a classification scenario with one class per task and multi-task continual reinforcement learning. We also use the conclusion from section \ref{sec:shortcomings} for the case of pre-trained models.

\textbf{- Learning from one class only}:
% one class per task a completely right situation
A classification task with only one class might look foolish, however, in a sequence of tasks with varying numbers of classes, it makes more sense and it seems evident that a CL algorithm should be able to handle this situation.
% noted by marginal vs conditional
Nevertheless, a classification neural network needs at least two classes to learn discriminative parameters.
% therefore la regu is useless
Hence, as mentioned in \cite{lesort2018marginal}, in a one-class task, the model can not learn meaningful parameters to preserve for later tasks. A regularization approach can then not protect any knowledge and enforce memorization.

\textbf{- Multi-task Continual Reinforcement Learning: }
Results from section~\ref{sec:shortcomings} can also be generalized to continual multi-tasks reinforcement learning scenarios \cite{Traore19DisCoRL}.
% remapping RL to Classif
In reinforcement learning, models do not predict classes but actions to realize.
% multi-task
Hence, a scenario where different tasks use disjoint sets of actions is equivalent to a class incremental learning scenario. The model has to predict different actions for each task.
In RL, learning is even harder since the supervision is only composed of sparse rewards and no direct supervision is usually given.
The same problem as for classification will then arise,
%will then arise
with the inability to learn a model that can correctly choose between actions from different tasks. The demonstration in section \ref{sub:demo} can directly apply since it is independent of the level of supervision.

\textbf{- Using pre-trained models for continual learning: }
We showed in Section \ref{sec:shortcomings} that, in a class incremental classification scenario, regularization methods are not sufficient to learn continually. 
In the case of a pre-trained classification model that we want to train on new classes without forgetting. We do not have the initial training data nor a \textit{regularization term} $\Omega$ to protect important weights. 
Following the demonstration in section \ref{sub:demo} and a fortiori without the regularization term, the modification of weights will lead the model to forget past knowledge. 
%

%%%%%%%%%%%%%%%%%%%%%%%%%%%%%%%%%%%%%%%%%%%%%%%%%%%%%%%%%%%%%%%%%%%%%%%%%%%%%%%%
%%%%%%%%%%%%%%%%%%%%%%%%%%%%%%%%%%%%%%%%%%%%%%%%%%%%%%%%%%%%%%%%%%%%%%%%%%%%%%%%
%%
%                                    Section
%%
%%%%%%%%%%%%%%%%%%%%%%%%%%%%%%%%%%%%%%%%%%%%%%%%%%%%%%%%%%%%%%%%%%%%%%%%%%%%%%%%
%%%%%%%%%%%%%%%%%%%%%%%%%%%%%%%%%%%%%%%%%%%%%%%%%%%%%%%%%%%%%%%%%%%%%%%%%%%%%%%%
\section{Discussion and Conclusion}
\label{sec:conclusion}

% regularization is famous but it has shortcomings
Regularization is a widespread approach for continual learning.
However, in this paper, we proved that those methods do not ensure continual learning in class-incremental scenarios.
They are unable to learn the discriminate inter-tasks classes. 
We conduct experiments on MNIST-Fellowship which corroborate with this conclusion.
We also illustrate this shortcoming with numerical and visual examples. 
One of the hypotheses we developed is that the issue might not be only catastrophic forgetting but also an inappropriate feature selection to solve past tasks. This \textit{features selection problem} might be one of the key problems of continual learning with catastrophic forgetting. 

% class-incremental scope
Our analysis is based on the class-incremental scenario.
Those scenarios allow us to measure the ability of algorithms to learn sequentially different classes. Being unable to deal with those scenarios implies that in a more complex learning environment, all sub-tasks assimilable to a class-incremental task might fail. 

% autonomy need
We have seen that the regularization shortcomings can be bypassed by using the task label for inference.
However, it makes the algorithm dependant on task annotation for prediction. 
We believe that training models that do not rely on task labels for inference is one of the main challenges of continual learning. Then, the use of replay-regularization coupled strategies could be an efficient alternative approach to protect past knowledge while learning continually.

\subsubsection*{Acknowledgments}
Thanks to Clément Pinard, Massimo Caccia, Lucas Caccia,Thomas George, Hugo Caselles-Dupré and Alexandre Chapoutot for interesting discussions for the realization of this paper. Thanks also to Vyshakh Palli Thazha for proofreading this article.

\bibliographystyle{splncs04}
\bibliography{continual,others}

\end{document}